\documentclass[10pt,twocolumn,letterpaper]{article}

\usepackage[applications]{wacv}      % To produce the REVIEW version for the applications track

\usepackage[T1]{fontenc}
\usepackage{amsmath,amssymb,amsfonts}
\usepackage{graphicx}
\usepackage{textcomp}
\usepackage{xcolor}
\usepackage{booktabs}
\usepackage{multirow}
\usepackage{url}
\usepackage{tikz}
\usetikzlibrary{calc}
\usetikzlibrary{arrows.meta}

\definecolor{wacvblue}{rgb}{0.21,0.49,0.74}
\usepackage[pagebackref,breaklinks,colorlinks,allcolors=wacvblue]{hyperref}

\def\wacvPaperID{1139} % *** Enter the WACV Paper ID here
\def\confName{WACV}
\def\confYear{2027}

\providecommand{\etal}{\emph{et al}.}   % no-op if wacv.sty already defines it

\begin{document}

\title{Facial Age Estimation for Age Fraud Detection in National ID Systems}

\author{Sharib Athar\footnotemark[1] \quad Arka Koner\footnotemark[1] \quad Chetan Naik\footnotemark[1] \quad Barada P. Sabut\footnotemark[1] \quad Tanusree Deb Barma\footnotemark[1] \\
Anoop M. Namboodiri\footnotemark[2] \quad Anil K. Jain\footnotemark[3]}

\maketitle

\begin{abstract}
Identity fraud during biometric enrollment and updates remains a major challenge for large-scale national identity systems. A common fraud vector is misrepresenting one's age to access age-restricted services or welfare schemes. In this work, we present SwinAge, a facial age estimation system designed for use within the Aadhaar biometric enrollment pipeline, to assist quality-check (QC) operators to flag potential age-related fraud. This is critical for a system like Aadhaar (the world's largest national identity programme), that holds about 1.5 billion unique identities, with 22.4 million new enrollments and 283 million updates in the last year. Building upon the SwinFace architecture with landmark-based similarity (warp affine) alignment, we train on a large in-house dataset of 1.45 million face images and evaluate on an independent, age-stratified test set of 283K images, both drawn from an ethnically diverse population of 716K unique subjects. We investigate three Aadhaar-specific operational thresholds (5, 18, and 60 years) and propose a deployment triage framework that flags suspected cases for manual review. Following NIST FATE, we report false acceptance/rejection rates (FAR/FRR) at each threshold rather than aggregate accuracy: at 1\% FAR the model achieves an FRR of 3\% ($<$5\,yrs), 0.4\% ($\geq$18\,yrs) and 11.0\% ($\geq$60\,yrs). SwinAge achieves a mean absolute error (MAE) of 2.94 years on the same test set, outperforming three zero-shot vision language models on all benchmarks, and improving the state-of-the-art on 5 out of 7 public benchmark datasets. We further report per-gender errors and distill lessons for national identity programs.
\end{abstract}

\footnotetext[1]{$^*$UIDAI, (\{techexe20.yp24, arka.koner, chetan.naik, hoe.dev-tc, tanusree.db\}@uidai.net.in)} 
\footnotetext[2]{$^\dagger$IIIT Hyderabad, (anoop@iiit.ac.in)}
\footnotetext[3]{$^\ddagger$Michigan State University, (jain@cse.msu.edu)}

\section{Introduction}
\label{sec:intro}

\begin{figure}[!t]
\centering
\includegraphics[width=0.72\columnwidth]{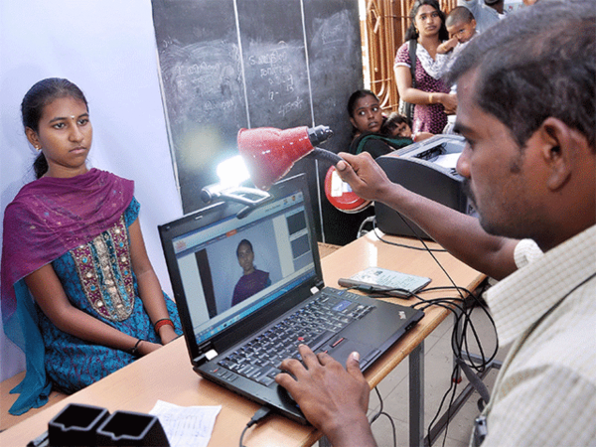}
\caption{Enrollment process for India's national ID system, called Aadhaar~\cite{b0}. The operator captures the applicant's facial photograph (as well as fingerprints and irises)  and demographic details, including date of birth. The declared age can be cross-checked against the facial age estimate to flag potential fraud.}
\label{fig:enrollment}
\end{figure}

National identity systems such as Aadhaar enroll millions of residents, collecting biometric and demographic data including name, date of birth, and facial photographs (Fig.~\ref{fig:enrollment}). In 2024--25 alone, the system processed 22.4 million new enrollments, 283 million updates, and 27.1 billion authentications~\cite{b30,b42}. The declared date of birth serves as the basis for age-dependent eligibility across numerous government and private services such as voting, driving licenses, employment, pension schemes, and child welfare programs. A recurring challenge during enrollment and demographic updates is age-related fraud: applicants intentionally misrepresenting their date of birth to access age-restricted services, enroll minors under adult identities, or exploit gaps in document verification.

A national performance audit by the Comptroller and Auditor General of India (CAG) on the National Social Assistance Programme revealed that over Rs.\,610 million (${\sim}$\$7.3M) was improperly disbursed to more than 95,000 ineligible beneficiaries, who failed to meet prescribed age thresholds~\cite{b25}. This included 57,394 individuals below age 60 receiving old-age pensions and 5,380 individuals below age 18 receiving disability benefits. A World Bank assessment of Thailand's social pension reports the same pattern of age-ineligible beneficiaries~\cite{b26}. Population-scale age claims are therefore not always reliable, and at Aadhaar's scale even a fraction of a percent of age misrepresentation can translate into hundreds of millions of rupees in annual benefit leakage. This motivates the use of automated facial age estimation as a triage of high-discrepancy cases for manual verification.

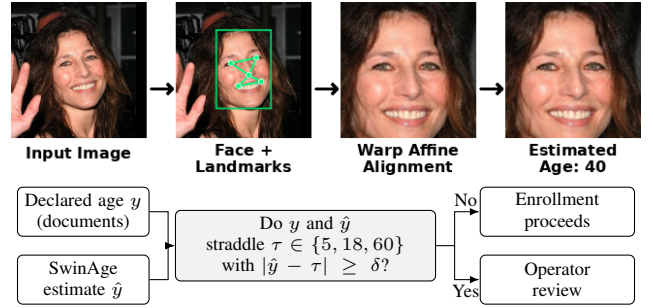
\begin{figure}[!t]
\centering
\includegraphics[width=\columnwidth]{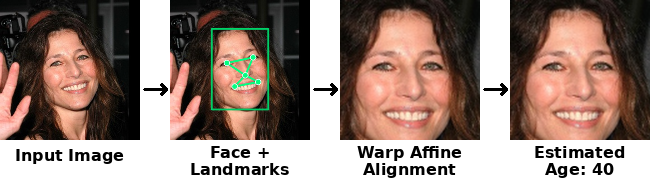}\\[3pt]
\begin{tikzpicture}[x=0.98mm, y=1mm, font=\scriptsize,   % x=0.98mm: was 1mm, overran \columnwidth by ~3pt
  n/.style={draw, rounded corners=2pt, align=center, inner sep=2pt, minimum height=6.5mm},
  ar/.style={-{Latex[length=1.3mm,width=1mm]}}]
\node[n, text width=16mm] (dec) at (0,0)  {Declared age $y$\\(documents)};
\node[n, text width=16mm] (est) at (0,-9) {SwinAge\\estimate $\hat{y}$};
\node[n, fill=black!5, text width=33mm] (cmp) at (30,-4.5)
  {Do $y$ and $\hat{y}$\\straddle $\tau\!\in\!\{5,18,60\}$\\with $|\hat{y}-\tau|\geq\delta$?};
\node[n, text width=19mm] (ok) at (64,0)  {Enrollment\\proceeds};
\node[n, text width=19mm] (rv) at (64,-9) {Operator\\review};
\draw[ar] (dec.east) -- ++(2.5,0) |- (cmp.west);
\draw[ar] (est.east) -- ++(2.5,0) |- (cmp.west);
\draw[ar] (cmp.east) -- ++(2.5,0) coordinate (mid) |- (ok.west);
\draw[ar] (mid) |- (rv.west);
\node[font=\scriptsize] at ($(mid)+(1.5, 6)$) {No};
\node[font=\scriptsize] at ($(mid)+(1.5,-6)$) {Yes};
\end{tikzpicture}
\caption{Aadhaar enrollment with SwinAge. Top: the operator's photograph, face and 5-landmark detection, warp affine alignment to the $112\times112$ template, and age estimation (public benchmark image). Bottom: the estimate is compared with the declared age at each threshold $\tau\!\in\!\{5,18,60\}$; if the two fall on opposite sides of $\tau$ (e.g., declared $<\!18$ but $\hat{y}\geq\!18$) and the estimate is at least $\delta$ away from $\tau$, the case is flagged for operator review (Sec.~\ref{sec:usecase_eval}).}
\label{fig:pipeline}
\end{figure}

Manual verification is subjective and exhaustive checks are infeasible on a national scale. The SwinAge system instead flags cases where the declared and estimated ages diverge significantly, prompting operator review (Fig.~\ref{fig:pipeline}). This does not replace document-based verification, but prioritizes cases that warrant manual scrutiny.

Three specific fraud cases motivate this work, each tied to an age threshold that governs access to services within the Aadhaar system:

\textit{1) Biometric bypass (declare age $<$5)}: Children below 5 are enrolled with a photograph only, with fingerprints and iris captured at the mandatory biometric updates at ages 5 and 15~\cite{b41}, and one may falsely claim a child's age to be $<$5 to enroll without biometric records.

\textit{2) Adult eligibility (declare age $\geq$18)}: Minors may falsely claim adulthood for voting, driving licenses, or employment eligibility.

\textit{3) Senior citizen benefits (declare age $\geq$60)}: Adults may falsely declare their age to be $\geq$60 to prematurely access pension schemes and welfare benefits.

\begin{figure}[!t]
\centering
\includegraphics[width=0.88\columnwidth]{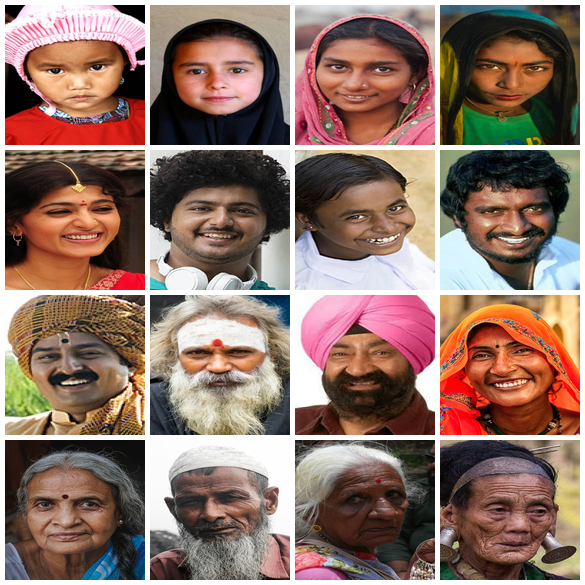}
\caption{Demographic diversity of the population served by Aadhaar. These examples from public sources are representative of the diversity in skin tone, facial features, hair styles, and aging patterns across different regions and ethnicities.}
\label{fig:diversity_panel}
\end{figure}

Despite significant progress in computer vision, facial age estimation remains challenging. Aging is a continuous and non-linear process, with varying rates of visual change across different life stages, and two people of the same chronological age may look significantly different due to genetics, lifestyle, and exposure to sun and other environmental factors (Fig.~\ref{fig:diversity_panel}). Moreover, building such a model is affected by dataset issues such as data imbalance (fewer samples at age extremes: $<$5 and 70+), label noise (data entry errors and approximate birth dates), and capture variability including lighting, pose, expression, and camera differences. NIST likewise reports that commercial age estimation accuracy depends strongly on sex, image quality, region of birth, and age itself~\cite{b33}.

In this work, we introduce SwinAge, a facial age estimation model built on the SwinFace architecture~\cite{b4} and tailored to the Aadhaar enrollment pipeline. To our knowledge, this is the first large-scale evaluation of age estimation on predominantly South Asian demographic data, a population characterized by significant diversity in skin tone, facial features, and aging patterns across regions and ethnicities. SwinAge is trained on 1.45M in-house images, evaluated on a 283K held-out test set (the full corpus comprises 1.75M images of 716,846 subjects), and cross-evaluated on seven public benchmark datasets. Our key contributions are:

\begin{itemize}
\item \textbf{Large-scale evaluation on a diverse real-life dataset} of 1.75 million images sampled from Aadhaar enrollment data, analyzing declared-versus-predicted age discrepancies to assess the ability to flag potential fraud cases in real-world scenarios.
\item \textbf{Detailed experimental analysis} across age groups and gender, and ablation of face alignment and backbone (frozen vs.\ fine-tuned).
\item \textbf{Comprehensive benchmarking on seven public datasets} (AgeDB~\cite{b6}, UTKFace~\cite{b7}, FGNet~\cite{b8}, APPA-REAL~\cite{b9}, CACD2000~\cite{b10}, MORPH2~\cite{b11}, AFAD~\cite{b32}) against published results from FaRL+MLP~\cite{b22}, AGMixer~\cite{b23}, SA-LDL~\cite{b24}, MiVOLO~\cite{b2}, FPAge~\cite{b5}, and DEX~\cite{b1}, and against three zero-shot vision-language models.
\item \textbf{Use-case-driven deployment framework} for three fraud scenarios (biometric bypass, adult eligibility, and senior benefits), with FAR/FRR at all three thresholds under an exact rule and a decision band, and the lessons these carry for identity programmes.
\end{itemize}

\section{Related Work}
\label{sec:related}

Facial age estimation has evolved from hand-crafted representations to deep learning-based approaches, such as regression~\cite{b15}, classification~\cite{b1}, label-distribution learning~\cite{b36,b37}, or ordinal regression~\cite{b32,b38}, with hybrids such as the mean--variance loss~\cite{b39}. DEX (Deep EXpectation)~\cite{b1} formulated apparent age estimation as classification followed by softmax expectation using VGG-16 ensembles.

Transformer-based architectures have subsequently become prominent. SwinFace~\cite{b4} employs a shared Swin Transformer~\cite{b3} backbone with task-specific subnets for recognition and age estimation, and MiVOLO~\cite{b2} introduced a multi-input transformer that combines facial and full-body data for joint age and gender estimation. FP-Age~\cite{b5} incorporates semantic face parsing with attention to focus on informative facial components under variations in pose, expression, and occlusion.

Recent works explore improved evaluation protocols, loss formulations, and auxiliary demographics. Paplham and Franc~\cite{b22} showed that factors like facial alignment, resolution, coverage, and pretraining substantially affect performance. AGMixer~\cite{b23} incorporates gender features with an improved ordinal loss, while SA-LDL~\cite{b24} introduces stage-wise adaptive label distribution learning to handle label ambiguity. Gender and race bias in age estimators has also been quantified~\cite{b19,b20,b40}; we therefore report per-gender and per-age-group errors.

Kuprashevich \etal~\cite{b31} evaluated multimodal LLMs such as ChatGPT, LLaVA-Next, and ShareGPT4V as general-purpose age estimators, while highlighting practical advantages of specialized models; we further include recent open-source vision-language models~\cite{b27,b28,b29} as zero-shot baselines. NIST's FATE age estimation track~\cite{b33} reports error at operational thresholds, and the ISO/IEC 27566-1 age assurance framework~\cite{b43} asks for false positive/negative rates, outcome error parity across demographic groups, and latency; we report against each. To our knowledge, no prior work evaluates age estimation on national-ID enrollment data, where the reference label is itself the record under audit.

\section{The SwinAge Model}
\label{sec:model}

We build upon the SwinFace architecture~\cite{b4}, which uses a Swin Transformer~\cite{b3} backbone to extract facial features at multiple resolutions through shifted-window self-attention. The backbone produces multi-scale local and global features that capture both fine-grained and holistic facial information.

After the backbone, features are passed into a task-specific head for age estimation. This branch contains a Feature Attention Module (FAM) with CBAM~\cite{b13} channel attention, followed by max-pooling and ReLU activation. Attention blocks suppress background variation and learn age-relevant features. Except for the backbone-finetuning ablation of Sec.~\ref{sec:alignment_effect}, the Swin Transformer backbone is kept frozen at the original pretrained SwinFace weights, and only the following layers are trained on our dataset. The architecture consists of:
\begin{itemize}
\item \textbf{Backbone}: Swin Transformer (frozen), producing multi-scale local and global features.
\item \textbf{Feature Attention Module (FAM)}: Conv $3\times3$ + CBAM channel attention + max pooling + ReLU.
\item \textbf{Task-Specific Subnet (TSS)}: MLP (Linear $\to$ ReLU $\to$ Dropout $\to$ Linear $\to$ ReLU $\to$ Dropout).
\item \textbf{Output Module (OM)}: Age (Linear $\to$ 1, regression).
\end{itemize}

SwinAge contains approximately 36.0M parameters in total, of which only 7.1M ($\sim$20\%) are trainable (the age estimation layers), while the remaining 28.9M parameters (Swin Transformer backbone) are frozen. This approach significantly reduces the training cost and the chance of overfitting, while retaining the rich facial representations learned during pretraining; unfreezing the backbone yields only a 0.03 year MAE gain (Sec.~\ref{sec:alignment_effect}).

\subsection{Face Detection and Alignment}
\label{sec:alignment}

\begin{figure}[!t]
\centering
\includegraphics[width=0.85\columnwidth]{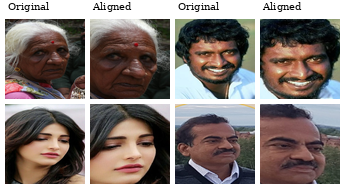}
\caption{Effect of warp affine alignment. Each pair shows the original face image (left) and aligned face (right) after similarity transformation using 5 landmarks. Side-facing poses in row 2 experience greater distortion, as the similarity transform assumes a near-frontal face.}
\label{fig:warping_effect}
\end{figure}

For detection, we employ an in-house-trained YOLOv8-nano~\cite{b12} face detector to detect faces and extract 5-point facial landmarks: the two eye centers, nose tip, and two mouth corners. The alignment step then normalizes face pose and geometry so that the backbone receives consistently aligned facial inputs: a similarity transformation is estimated to align the 5 detected landmarks to the canonical $112\times112$ ArcFace face template~\cite{b14}, correcting for in-plane rotation, translation, and scale (``warp affine'' hereafter). Matching the backbone's pretraining alignment is what lets its frozen features transfer (Sec.~\ref{sec:alignment_effect}). Fig.~\ref{fig:warping_effect} illustrates the effect. An alternative is tanh-polar warping~\cite{b34}, used by FP-Age~\cite{b5}, which transforms the face into a tanh-polar coordinate system emphasizing central features.

\subsection{Loss Function}
\label{sec:loss}

For age prediction, we use a hybrid loss that smoothly transitions during training. In early training, it behaves like a squared error (MSE) loss, encouraging stable optimization. As training progresses, it gradually shifts towards a robust Gaussian (Welsch/Leclerc) loss~\cite{b35,b15}, reducing the influence of large outliers from noisy labels:
\begin{equation}
\mathcal{L}_{age} = (1 - \lambda) \frac{(y - \hat{y})^2}{2} + \lambda \left(1 - e^{-\frac{(y - \hat{y})^2}{2\sigma^2}}\right)
\label{eq:loss}
\end{equation}
where $y$ and $\hat{y}$ are the true and predicted age, $\lambda = \frac{t}{T}$ is the progress ratio (current step $t$ over total steps $T$), and $\sigma = 3.0$ is a fixed scale parameter controlling the penalty for error during the robust regime. This value was selected using a validation set and held constant across all experiments.

\subsection{Training Configuration}
\label{sec:training}

SwinAge is initialized from the pretrained SwinFace weights~\cite{b4}, which include both the Swin Transformer backbone (pretrained on face recognition via ArcFace loss~\cite{b14}) and the task-specific head (FAM, TSS, OM). During training, the backbone is kept frozen and only the task-specific head is updated. Aligned inputs are normalized to $[-1,1]$ and augmented with horizontal flips, RandAugment~\cite{b44} ($n{=}2$, $m{=}9$), and random erasing~\cite{b45} ($p{=}0.25$). We use the AdamW~\cite{b16} optimizer with weight decay $0.05$, a linear warmup of 8000 steps, and a peak learning rate of $5 \times 10^{-3}$. The model is trained for 80,000 steps with a total batch size of 512 on two Tesla V100-32GB GPUs using FP16 mixed precision, taking approximately 60~h.

\section{Datasets}
\label{sec:datasets}

\subsection{In-House Dataset}
\label{sec:inhouse_data}

Our primary dataset consists of 1.75 million face images captured during Aadhaar biometric enrollment, representing 716,846 unique subjects, where the demography is predominantly South Asian. Its use is governed by the Aadhaar Act~\cite{b17} and the Digital Personal Data Protection (DPDP) Act~\cite{b18}, which prohibit publication or sharing of any biometric data but allow use of resident data to improve service delivery and fraud prevention. This study uses a fully anonymized set under that provision, with all personally identifiable information (name, Aadhaar number, address, contact details) removed prior to use; images of minors (0--18) were subject to the same controls. No raw images from the database are published or shared in this paper. The dataset has significant diversity in skin tone, facial features, hair styles, and aging patterns across regions and ethnicities (Fig.~\ref{fig:diversity_panel}).

The dataset is split into training (1,451,294 images), validation (16,941 images), and test (283,515 images) sets, totaling 1,751,750 images after preprocessing. Subject-level splitting ensures no identity leakage between splits; the test split covers 192,096 unique subjects (97,199 male, 94,871 female, and 26 transgender). The test set is age-stratified for per-group analysis, so the overall MAE reflects a balanced rather than the real-world age distribution. The age label throughout is the \textit{declared} date of birth on record, so the reported MAE reflects the error with the declared age rather than chronological age.

\subsection{Public Datasets}
\label{sec:public_data}

\begin{figure*}[!t]
\centering
\includegraphics[width=\textwidth]{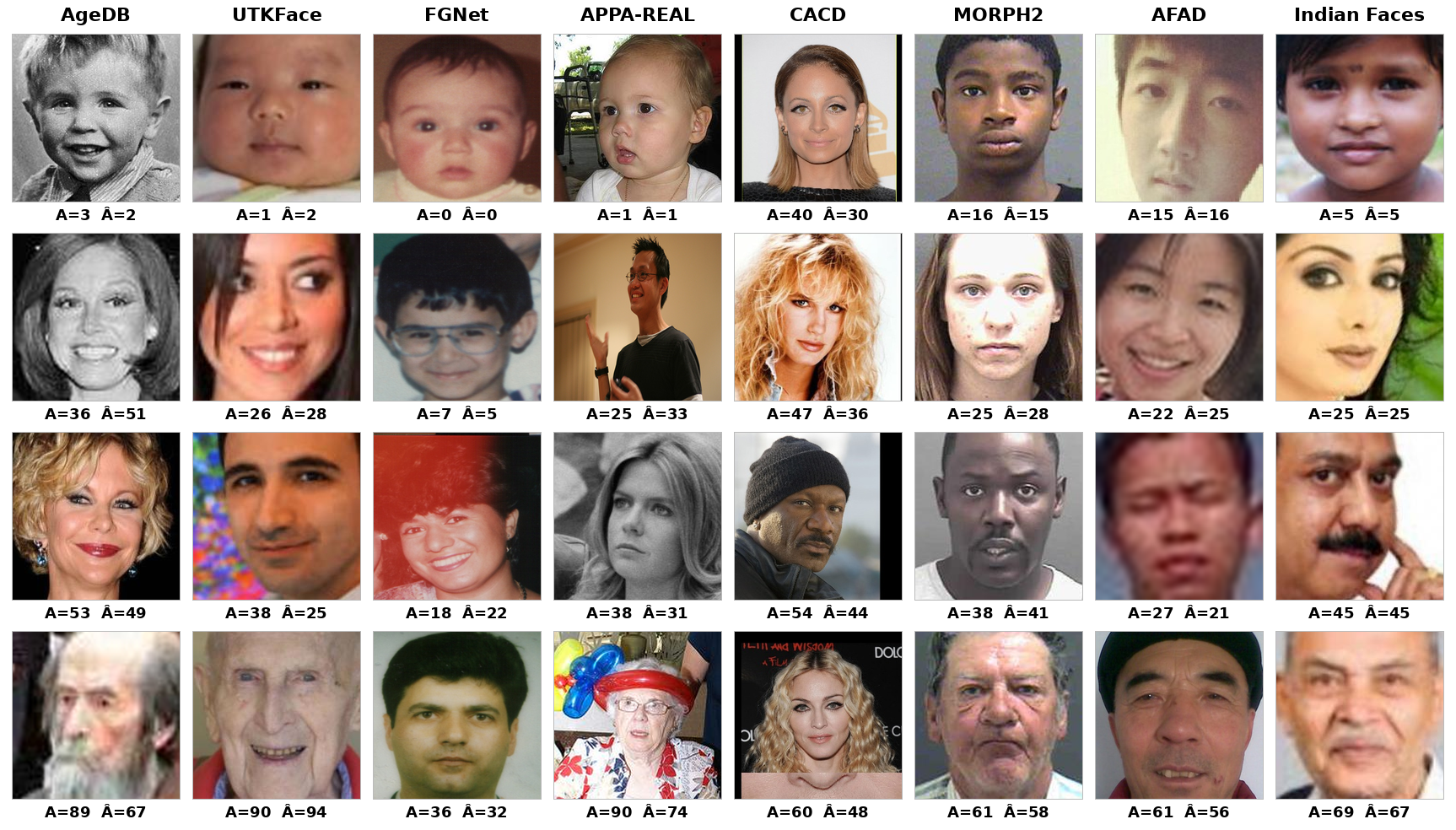}
\caption{Age-estimation results on seven public datasets and a South Asian demographic subset. All images are from public sources; no National ID image is shown. Below each image: $A$ = true age, $\hat{A}$ = predicted age.}
\label{fig:collage}
\end{figure*}

We evaluate SwinAge on six widely used public age benchmarks plus AFAD for intra-dataset comparison (Table~\ref{tab:datasets}). These datasets vary significantly in size, age range, image quality, and capture conditions. Image counts may differ slightly from the original releases due to filtering of images where face detection or landmark extraction failed. Fig.~\ref{fig:collage} shows sample images from all seven public datasets plus a South Asian demographic subset drawn from public sources.

For combined ``All'' dataset experiments, we finetune the in-house trained model on a mixture of public datasets alongside the in-house data. For each public dataset, we use the provided train/test split when available; otherwise we partition randomly. Datasets with more than 20,000 training samples are subsampled to 20,000 to prevent dominance. Training and test identities are kept disjoint wherever identity labels are available.

\begin{table}[!t]
\caption{Summary of evaluation datasets}
\label{tab:datasets}
\centering
\setlength{\tabcolsep}{3.5pt}   % was 4pt: overran \columnwidth by ~3pt
\renewcommand{\arraystretch}{1.05}
\small
\begin{tabular}{lrrrl}
\toprule
\textbf{Dataset} & \textbf{Images} & \textbf{Ages} & \textbf{Subjects} & \textbf{Source} \\
\midrule
AgeDB \cite{b6} & 16,463 & 3--101 & 440 & celebrity \\
UTKFace \cite{b7} & 22,666 & 0--116 & N/A & in-the-wild \\
FGNet \cite{b8} & 984 & 0--69 & 82 & longitudinal \\
APPA-REAL \cite{b9} & 7,519 & 1--100 & 7000+ & in-the-wild \\
CACD2000 \cite{b10} & 154,900 & 14--62 & 2,000 & celebrity \\
MORPH2 \cite{b11} & 50,013 & 16--77 & 19,033 & mugshot \\
AFAD \cite{b32}$^{\mathrm{b}}$ & 165,515 & 15--75 & N/A & Asian, web \\
National ID$^{\mathrm{a}}$ & 1,751,750 & 0--75 & 716,846 & enrollment \\
\bottomrule
\end{tabular}
\begin{flushleft}
\footnotesize
$^{\mathrm{a}}$Privacy law prevents the display of any image from the National ID database. $^{\mathrm{b}}$Intra-dataset evaluation only; not in joint (``All'') training.
\end{flushleft}
\end{table}

\section{Experiments and Results}
\label{sec:experiments}

\subsection{Experimental Setup}
\label{sec:setup}

We evaluate seven configurations (Table~\ref{tab:overall}): (1) the original pretrained SwinFace weights evaluated directly on in-house data, with no additional training; the FP-Age architecture~\cite{b5} (ResNet50 + face parser) trained on in-house data (2) without face alignment, (3) with tanh-polar warping as in the original paper, and (4) with landmark-based warp affine alignment; (5) SwinFace with frozen backbone, trained only on the task head, without alignment; (6) SwinFace with tanh-polar warping; and (7) SwinAge (proposed), i.e.\ (5) with warp affine alignment. Models 2--7 are trained on the same 1.45M in-house dataset with identical training configurations; the only variables are the architecture and preprocessing, enabling a controlled ablation of these design choices.

\subsection{Use-Case-Driven Evaluation}
\label{sec:usecase_eval}
We first evaluate SwinAge's ability to correctly classify subjects 
relative to each of the three service-eligibility thresholds 
($\tau \in \{5, 18, 60\}$) introduced in Sec.~\ref{sec:intro}. 
The privileged side differs by threshold: for $\tau{=}5$ a fraudster 
claims to be \emph{below} $\tau$ (biometric bypass), while for 
$\tau \in \{18, 60\}$ a fraudster claims to be \emph{above} $\tau$ 
(adult rights, senior benefits); the fraud-relevant error is FAR at 
all three thresholds. Letting $y$ denote true age, $\hat{y}$ predicted 
age and $\delta$ margin:

{\begingroup
\footnotesize
\setlength{\arraycolsep}{2pt}
\begin{align*}
\text{FAR}_5      &= P(\hat{y} < 5{-}\delta \mid y \geq 5),       & \text{FRR}_5      &= P(\hat{y} \geq 5{+}\delta \mid y < 5),       \\
\text{FAR}_{18}   &= P(\hat{y} \geq 18{+}\delta \mid y < 18),     & \text{FRR}_{18}   &= P(\hat{y} < 18{-}\delta \mid y \geq 18),     \\
\text{FAR}_{60}   &= P(\hat{y} \geq 60{+}\delta \mid y < 60),     & \text{FRR}_{60}   &= P(\hat{y} < 60{-}\delta \mid y \geq 60)
\end{align*}
\endgroup}

Both are computed from continuous predicted and true ages, so that errors near the threshold are not absorbed by a wide bin; rates within a NIST-style restricted challenge window would be higher than these whole-population rates.

% With an exact rule ($\delta{=}0$) every small regression error at the boundary becomes a decision error. We therefore also evaluate a decision band of margin $\delta$: a subject is flagged as above $\tau$ only when $\hat{y} \geq \tau+\delta$, flagged as below $\tau$ only when $\hat{y} < \tau-\delta$, and otherwise deferred to document verification, in which case the record proceeds with no flag. FAR and FRR then count confident errors only, over the 91--99\% of records the band decides (1.3\%, 9.1\%, and 8.6\% deferred at $\tau{=}5$, 18, and 60). $\delta$ is the smallest margin giving FAR $\leq 1\%$ for each threshold; the reported values were read off the test set, so the band columns are an operating point rather than an independently validated one, and the exact-rule columns are the unbiased reference. Table~\ref{tab:threshold} gives both settings. At $\tau{=}60$ the exact rule rejects 21.4\% of genuine seniors and the band still 11.0\%, so the $\geq$60 FRR is the binding constraint on deployment, while its FAR remains low, so that few non-seniors would be classified as eligible for senior benefits.

With an exact rule ($\delta{=}0$) every small regression error at the boundary becomes a decision error. We therefore also evaluate a decision band of margin $\delta$: a subject is flagged as above $\tau$ only when $\hat{y} \geq \tau+\delta$, flagged as below $\tau$ only when $\hat{y} < \tau-\delta$, and otherwise deferred to document verification. FAR and FRR then count confident errors only, over the 91--99\% of records the band decides.

We report two band settings. The first sets $\delta$ to the smallest value giving FAR $\leq 1\%$, read off the test set; these columns are an operating point rather than an independently validated one, and the exact-rule columns are the unbiased reference. The second uses $\delta{=}0.1\tau$, a model-agnostic fixed fraction that coincidentally aligns with the model's per-threshold MAE (0.75 at $\tau{=}5$, 2.07 at $\tau{=}18$, 3.76 at $\tau{=}60$), providing a deployment-ready rule requiring no held-out calibration. Table~\ref{tab:threshold} gives all three settings. At $\tau{=}60$ the exact rule rejects 21.4\% of genuine seniors; the $\geq$60 FRR remains the binding deployment constraint regardless of band setting, while FAR stays low at all thresholds.

% \begin{table}[!t]
% \caption{Threshold performance on the in-house test set (283,515 images), computed from continuous ages. With the decision band $\delta$, FAR/FRR count confident errors only.}
% \label{tab:threshold}
% \centering
% \setlength{\tabcolsep}{4pt}
% \small
% \begin{tabular}{lcc|cc}
% \toprule
%  & \multicolumn{2}{c|}{Exact ($\delta = 0$)} & \multicolumn{2}{c}{Decision band @ 1\% FAR} \\
% \cmidrule(lr){2-3} \cmidrule(lr){4-5}
% $\tau$ & FAR & FRR & \hspace{15pt}$\delta$ (yrs) & FRR \\
% \midrule
% 5  & 1.47\% & 4.95\%  & \hspace{18pt}0.3 & 3.00\% \\
% 18 & 5.67\% & 2.35\%  & \hspace{18pt}2.9 & 0.38\% \\
% 60 & 2.59\% & 21.39\% & \hspace{18pt}2.7 & 11.04\% \\
% \bottomrule
% \multicolumn{5}{l}{\footnotesize $^*$FAR and FRR are defined in Sec.~\ref{sec:usecase_eval}.}
% \end{tabular}
% \end{table}

\begin{table}[!t]
\caption{Threshold performance on the in-house test set (283,515 images), computed from continuous ages. With the decision band $\delta$ (in years), FAR/FRR$^*$ count confident errors only.}
\label{tab:threshold}
\centering
\setlength{\tabcolsep}{4pt}
\small
\begin{tabular}{lcc|ccc|cc}
\toprule
 & \multicolumn{2}{c|}{Exact ($\delta{=}0$)} 
 & \multicolumn{3}{c|}{$\delta{=}0.1\tau$} 
 & \multicolumn{2}{c}{$\delta$ @ 1\% FAR} \\
\cmidrule(lr){2-3}\cmidrule(lr){4-6}\cmidrule(lr){7-8}
$\tau$ & FAR & FRR & $\delta$ & FAR & FRR & $\delta$ & FRR \\
\midrule
5  & 1.47\% & 4.95\%  & 0.5 & 0.77\% & 2.13\% & 0.3 & 3.00\% \\
18 & 5.67\% & 2.35\%  & 1.8 & 2.15\% & 0.78\% & 2.9 & 0.38\% \\
60 & 2.59\% & 21.39\% & 6.0 & 0.17\% & 3.59\% & 2.7 & 11.04\% \\
\bottomrule
\multicolumn{8}{l}{\footnotesize $^*$FAR and FRR defined in Sec.~\ref{sec:usecase_eval}.}
\end{tabular}
\end{table}

% ---- The original tolerance-based confusion matrix (confusion_matrix_tolerance.pdf)
% is omitted: its bin tolerances were derived from the model's own MAE, which is
% the construction Table 2 replaces, and a reviewer would ask why both appear.
% Restore here only as a descriptive figure with its tolerances stated in the
% caption.

\textit{Deployment triage}: SwinAge makes no autonomous decisions. For each 
enrollment, the predicted age is compared against the declared age; if the 
estimate crosses a threshold by more than $\delta$, the case is flagged for 
manual operator review. The model serves only as an independent cross-check and the 
declared age remains as the primary source of truth. Reporting FAR/FRR at 
operational thresholds follows the NIST FATE protocol~\cite{b33}.

\subsection{In-House Dataset Evaluation}
\label{sec:inhouse_results}

\subsubsection{Ablation Study: Architecture and Alignment}

Table~\ref{tab:overall} presents the study results on the in-house test set, systematically varying the backbone architecture and face alignment method while keeping all other training settings fixed. SwinAge achieves the best overall MAE of 2.94.

\begin{table}[!t]
\caption{In-house test set results (MAE, years). Models 2--7 are trained with identical settings; only the architecture and alignment method vary. Model 1 uses pretrained weights without retraining.}
\label{tab:overall}
\centering
\setlength{\tabcolsep}{6pt}
\small
\begin{tabular}{lc}
\toprule
\textbf{Model} & \textbf{MAE} \\
\midrule
SwinFace \cite{b4} (pretrained)  & 6.99 \\
FPAge \cite{b5} (no warping)     & 3.51 \\
FPAge (tanh-polar)               & 3.06 \\
FPAge (warp affine)              & 3.16 \\
SwinFace (no alignment)          & 3.77 \\
SwinFace (tanh-polar)            & 3.88 \\
\textbf{SwinAge (proposed)}      & \textbf{2.94} \\
\bottomrule
\end{tabular}
\end{table}

\subsubsection{Per-Age-Group and Gender Analysis}
\label{sec:age_gender}

\begin{figure}[!t]
\centering
\includegraphics[width=0.85\columnwidth]{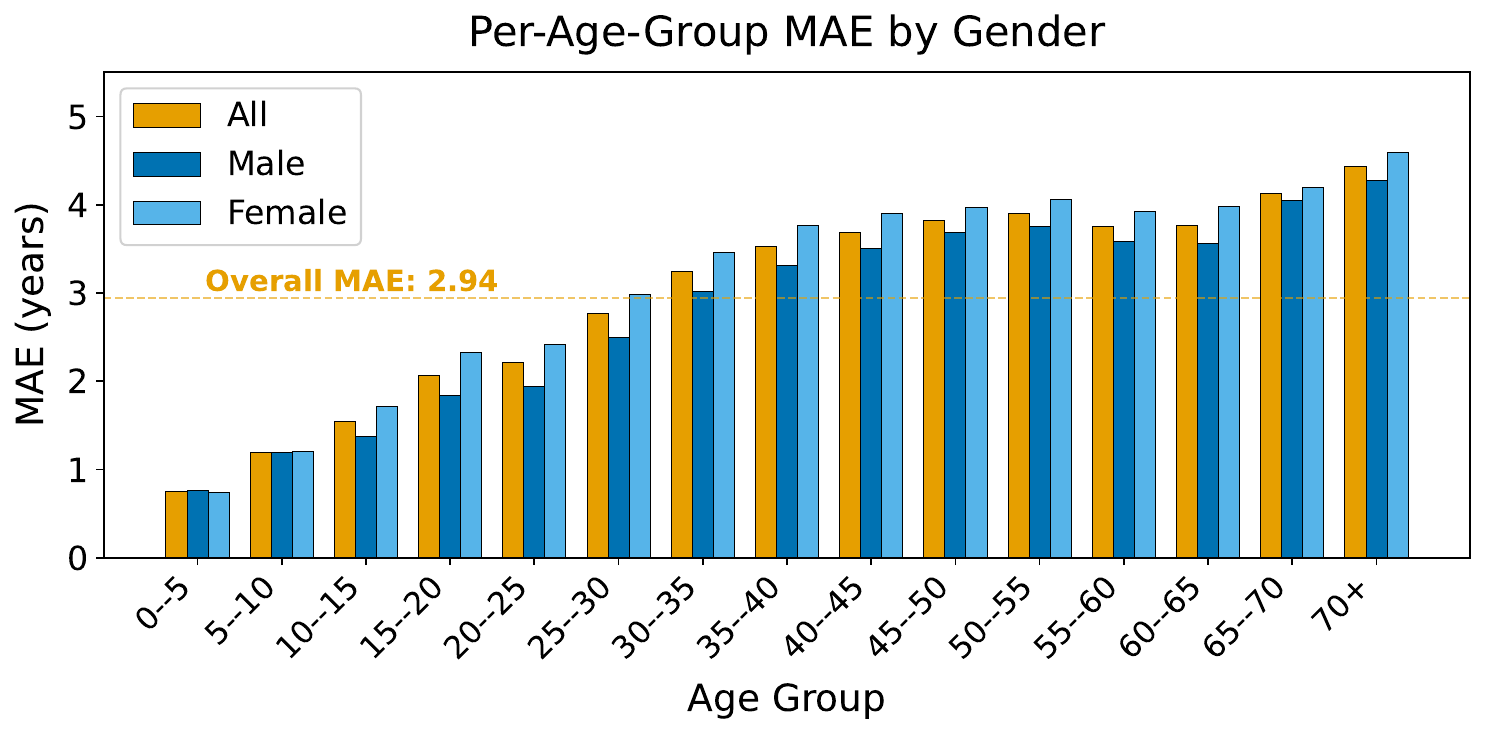}
\caption{Per-age-group MAE for SwinAge, stratified by gender. Orange bars show overall MAE, navy blue shows male, and sky blue shows female. The dashed line indicates the overall MAE of 2.94.}
\label{fig:age_group_mae}
\end{figure}

Fig.~\ref{fig:age_group_mae} presents detailed per-age-group results for SwinAge, including MAE stratified by gender. Age estimation is more accurate for male subjects (MAE 2.79) than female subjects (MAE 3.10). This gender gap is consistent across most age groups, with the largest differences observed in the 15--20 range. This aligns with prior work suggesting that female faces are more challenging, potentially due to cosmetics and styling that reduce apparent aging cues~\cite{b19,b20}, and with the gender effect NIST reports across commercial algorithms~\cite{b33}. Because what a resident experiences is being routed to manual review rather than the 0.3-year MAE gap across genders, demographic parity should be audited at the decision level of Table~\ref{tab:threshold}, with the $\geq$60 FRR under the most scrutiny; the 42 transgender images (26 subjects) are too few to report separately.

\begin{table*}[!t]
\caption{Age estimation comparison (MAE) of existing models on public datasets, each fine-tuned on those datasets (Intra). The ``All'' column gives the MAE of a single model trained on the combined training splits of all datasets except AFAD, evaluated per dataset. VLMs are evaluated zero-shot on cropped face images. \textbf{Best result per row is shown in bold}. -- indicates results not reported in the respective papers. The FPAge in-house entry is our own trained model (Table~\ref{tab:overall}); all other baseline entries are published numbers under their original protocols.}
\label{tab:intra_comparison}
\centering
\setlength{\tabcolsep}{4pt}
\small
\begin{tabular}{lccccccccccc}
\toprule
\multirow{3}{*}{\textbf{Dataset}} & \multicolumn{2}{c}{\textbf{SwinAge (ours)}} & \multicolumn{6}{c}{\textbf{Specialized age estimators}} & \multicolumn{3}{c}{\textbf{Zero-shot VLMs}} \\
\cmidrule(lr){2-3} \cmidrule(lr){4-9} \cmidrule(lr){10-12}
& & & SA-LDL & AGMixer & FaRL+MLP & MiVOLO & FPAge & DEX & Nemotron & Qwen3-VL & Gemma \\
& All & Intra & \cite{b24} & \cite{b23} & \cite{b22} & \cite{b2} & \cite{b5} & \cite{b1} & \cite{b27} & \cite{b28} & \cite{b29} \\
\midrule
UTKFace     & 4.34 & 4.16 & 4.45 & \textbf{3.67} & 3.87 & 3.99 & --   & --   & 5.38 & 6.05 & 5.03 \\
AFAD        & --$^*$ & \textbf{2.88} & --   & 2.90 & 3.12 & --   & --   & --   & 4.65 & 5.85 & 5.39 \\
FGNet & 2.89 & \textbf{2.15} & --   & 2.16 & --   & --   & 5.60 & --   & 5.16 & 4.51 & 4.19 \\
CACD2000    & 4.55 & 3.62 & 4.21 & \textbf{3.28} & 3.96 & --   & 4.33 & --   & 7.24 & 6.71 & 6.30 \\
AgeDB       & 4.53 & \textbf{4.21} & --   & 5.46 & 5.64 & 5.55 & --   & --   & 7.42 & 7.59 & 6.65 \\
APPA-REAL   & 4.75 & \textbf{4.37} & --   & --   & --   & --   & --   & 6.26 & 6.12 & 5.53 & 5.08 \\
MORPH2      & 2.15 & \textbf{1.57} & 1.75 & --   & --   & --   & 1.90 & --   & 5.45 & 6.12 & 4.53 \\
\midrule
In-house    & 3.47 & \textbf{2.94} & --   & --   & --   & --   & 3.06 & --   & 4.78 & 5.14 & 4.16 \\
\bottomrule
\end{tabular}
\begin{flushleft}
\footnotesize
$^*$AFAD is used for intra-dataset evaluation only and not in the combined ``All'' training.
\end{flushleft}
\end{table*}

\subsection{Public Dataset Evaluation}
\label{sec:cross_dataset}

We evaluate SwinAge under two finetuning settings. ``Intra'' denotes evaluation within each individual dataset, where the model is fine-tuned on the training split of that dataset and evaluated on its corresponding held-out test split. ``All'' denotes evaluation using a model fine-tuned on the combined training splits of six public datasets (AFAD was excluded) and the in-house dataset, with results reported separately for each constituent dataset's test split.\footnote{Model weights fine-tuned on the public datasets will be released upon acceptance of this paper.} As Table~\ref{tab:intra_comparison} shows, joint finetuning yields a single model that remains competitive on every dataset but is worse than dataset-specific finetuning on each, including the in-house set, likely due to domain conflict between the diverse public datasets and the controlled in-house capture conditions.

Table~\ref{tab:intra_comparison} compares SwinAge's intra-dataset results against published models, ordered by recency: SA-LDL~\cite{b24}, AGMixer~\cite{b23}, FaRL+MLP~\cite{b22}, MiVOLO~\cite{b2}, FPAge~\cite{b5}, and DEX~\cite{b1}. We additionally compare against three general-purpose vision-language models (Nemotron-Nano-12B~\cite{b27}, Qwen3-VL-235B~\cite{b28}, and Gemma-4-31B~\cite{b29}), evaluated zero-shot on cropped face images with a simple prompt. Despite no task-specific training, the VLMs are a strong baseline; SwinAge nevertheless outperforms all three on every dataset (e.g., 2.94 vs.\ 4.16--5.14 on in-house data), and matches or improves on five of the seven public datasets, two of them (AFAD, FGNet) by 0.01--0.02 years. Comparisons with the specialized estimators should be read with care: protocol differences alone can move MAE by more than the gaps at issue~\cite{b22}.

\section{Discussion}
\label{sec:discussion}

\subsection{Effect of Face Alignment}
\label{sec:alignment_effect}

\textit{1) Alignment substantially improves performance}: FPAge without warping (MAE 3.51) is substantially worse than both tanh-polar (3.06) and warp affine (3.16) variants. Similarly, SwinAge without alignment (MAE 3.77) performs substantially worse than with warp affine (MAE 2.94), showing that alignment preprocessing is a primary driver of improvement when using frozen pretrained features.

\textit{2) Alignment must match the architecture and capture conditions}: Warp affine achieves the best result for SwinAge (MAE 2.94), while tanh-polar \textit{degrades} performance below even no alignment (3.88 vs.\ 3.77), as its background-suppression design is redundant for our controlled, near-frontal captures. On FPAge, tanh-polar slightly outperforms warp affine (3.06 vs.\ 3.16), consistent with its face-parsing design intent.

\textit{3) Backbone finetuning yields marginal gains}: Unfreezing the backbone at a conservative learning rate ($5 \times 10^{-5}$, 0.01$\times$ the head LR) produced only a 0.03-year improvement in test MAE (2.91 vs.\ 2.94), too small to claim as a difference, and validation performance plateaued early (at step 10,000) with no further improvement over the remaining 70,000 steps.

\subsection{Age-Group-Specific Analysis}
\label{sec:age_analysis}

Performance varies significantly across age groups (Fig.~\ref{fig:age_group_mae}). Children (0--5) are the easiest to estimate (MAE 0.75), as infants and toddlers have distinctive facial features that change rapidly. The hardest range is 45--70+ (MAE $\sim$3.75--4.44), where aging changes are subtle and gradual and the cumulative effects of genetics, lifestyle, and sun exposure make people of the same age look increasingly different. This profile is why $\delta$ is chosen per threshold in Sec.~\ref{sec:usecase_eval}; part of the 60+ degradation is a range effect, since the corpus is capped at 75 and the predictions compress downward at the top.

\subsubsection{Regression-to-Mean: A Per-Dataset Bias Analysis}
\label{sec:bias_by_dataset}

\begin{figure*}[!t]
\centering
\includegraphics[width=\textwidth]{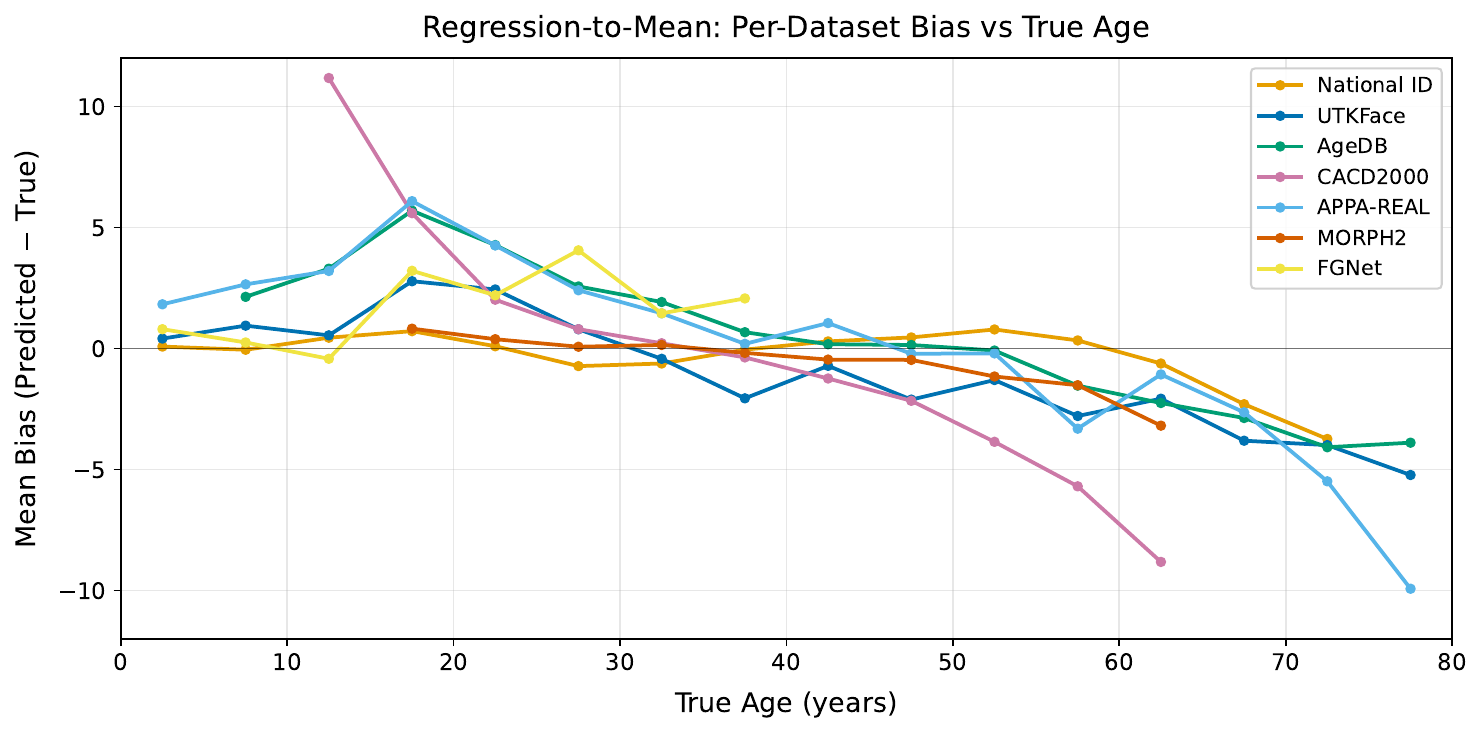}
\caption{Mean bias (predicted~$-$~true age) versus true age, in 5-year bins, for each dataset (Intra setting). All datasets exhibit the regression-to-mean pattern: positive bias (overestimation) for young subjects, negative bias (underestimation) for older subjects, with a crossover near 30--40~years. CACD2000 shows the most extreme swing ($+$5.8 at age ${<}20$ to $-$8.8 at age ${\geq}60$), reflecting apparent-age bias from celebrity styling. MORPH2 and National ID are the most balanced, consistent with their controlled capture conditions. Note: the bias values here use broader age bins ($<$20, $\geq$60) than the narrower bins cited in Sec.~6.2 of the main paper.}
\label{fig:bias_by_dataset}
\end{figure*}

A consistent regression-to-mean pattern is prevalent across all datasets: the model overestimates younger subjects and underestimates older ones, with a crossover near age 30--40. Fig.~\ref{fig:bias_by_dataset} visualizes this across all datasets by plotting the mean prediction error (predicted~$-$~true age) in 5-year bins of true age.

On UTKFace, children aged 0--5 are estimated near-perfectly (MAE 0.79) while subjects aged 70+ are underpredicted by $-$5.0 years; on AgeDB, bias shifts from $+$3.3 to $-$4.8 years across the same range despite near-zero overall bias ($-$0.08). CACD2000 shows the same split ($+$5.0 vs.\ $-$4.9 years), reflecting apparent-age bias from celebrity styling; for APPA-REAL, MAE drops from 4.75 to 4.22 against apparent rather than chronological labels, confirming the model tracks perceived age. MORPH2 performs best (MAE 1.57--2.15) owing to its frontal mugshot format, though its 85\% male composition yields a gender gap (MAE 2.05 vs.\ 2.74). Across all datasets, prediction variance is compressed to 90--96\% of true age variance, directly motivating the per-threshold $\delta$ calibration in  Sec.~\ref{sec:usecase_eval}, as the $\geq$60 FRR is driven by systematic underprediction of older faces.

Table~\ref{tab:bias_summary} quantifies the pattern. The variance compression column reports the ratio of predicted-age variance to true-age variance; values below 100\% indicate that the model's predictions are compressed toward the mean relative to the true distribution, which is the direct cause of the regression-to-mean bias. The main paper cites a representative range of 90--96\%; the full per-dataset breakdown is given here, with the note that National ID (97.3\%) and MORPH2 (97.6\%) slightly exceed this range, while AgeDB (82.4\%) falls below it due to its wider age span (3--101) amplifying edge effects.

\begin{table}[!t]
\caption{Per-dataset bias summary (Intra setting). ``$<$20 bias'' and ``$\geq$60 bias'' are the mean prediction errors for subjects below 20 and at or above 60~years, respectively. These age-bin values differ from the finer-grained figures cited in Sec.~6.2 of the main paper (e.g., CACD2000 $+$5.0/$-$4.9 for ages 14--20 and 50--62), which use narrower bins. ``Var.\ comp.'' is the ratio of predicted-age variance to true-age variance.}
\label{tab:bias_summary}
\centering
\setlength{\tabcolsep}{4pt}
\small
\begin{tabular}{lcccc}
\toprule
\textbf{Dataset} & \textbf{Overall} & \textbf{$<$20 bias} & \textbf{$\geq$60 bias} & \textbf{Var.\ comp.} \\
\midrule
National ID  & $-$0.33 & $+$0.30  & $-$2.23  & 97.3\% \\
UTKFace      & $-$0.47 & $+$0.96  & $-$3.88  & 93.1\% \\
AgeDB        & $-$0.08 & $+$4.77  & $-$3.74  & 82.4\% \\
CACD2000     & $-$0.91 & $+$5.78  & $-$8.82  & 91.8\% \\
APPA-REAL    & $+$1.28 & $+$3.81  & $-$4.81  & 86.0\% \\
MORPH2       & $-$0.03 & $+$0.82  & $-$3.51  & 97.6\% \\
FGNet        & $+$1.34 & $+$0.97  & $-$14.11 & 111.5\% \\
\bottomrule
\end{tabular}
\end{table}

The FGNet variance compression exceeds 100\% because its 60--65 bin contains a single heavily mislabeled image (MAE 14.1), inflating the predicted-age variance; with only 82 subjects, FGNet is the most sensitive to individual outliers. Excluding this bin, the compression drops to 95.2\%, consistent with the other datasets.

%% ================================================================
\subsection{Qualitative Failure Analysis}
\label{sec:failures}

\begin{figure*}[!t]
\centering
\includegraphics[width=\textwidth]{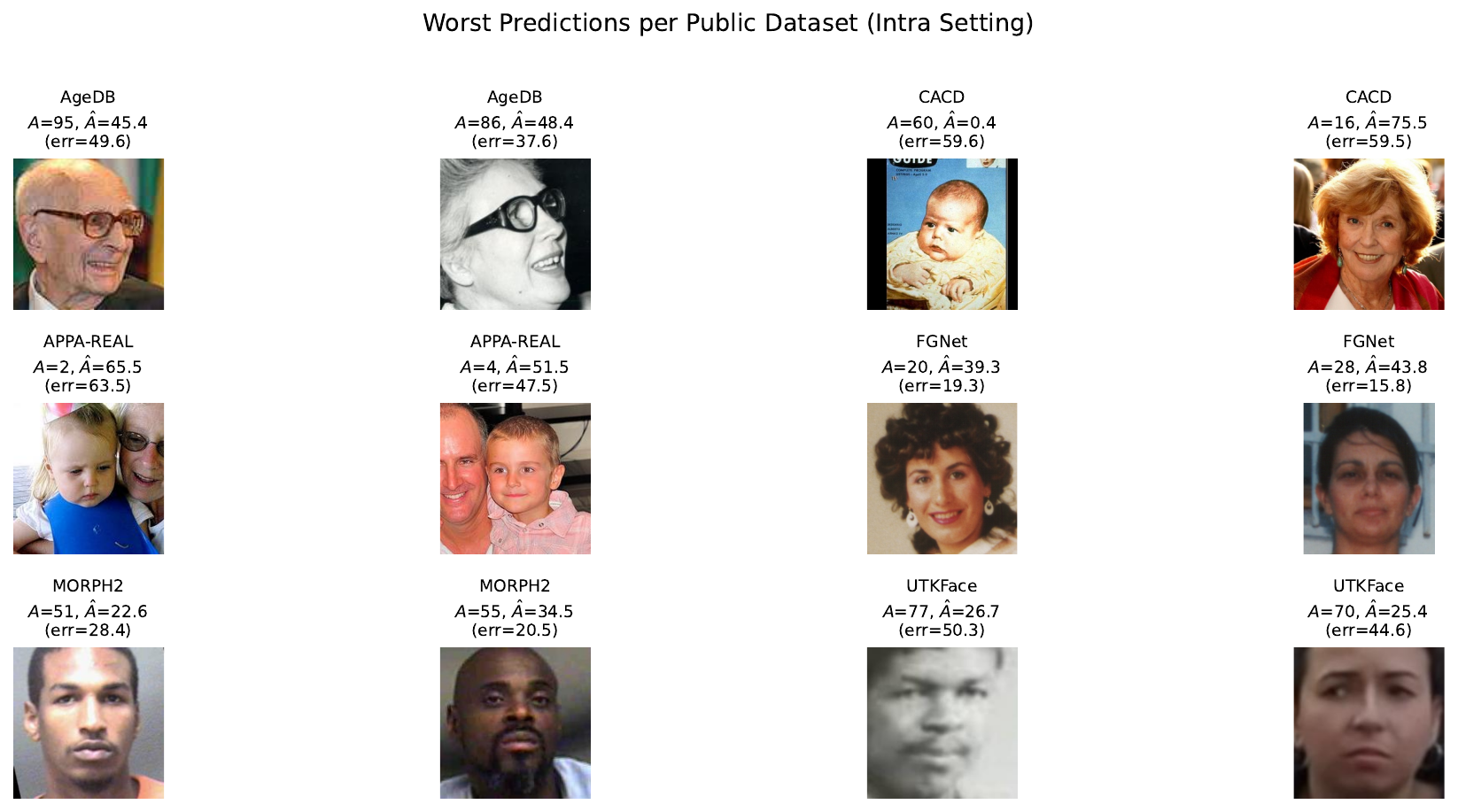}
\caption{Worst predictions per public dataset (Intra setting). Below each image: $A$ = true age, $\hat{A}$ = predicted age. Failure patterns include: (a)~extreme age combined with atypical appearance (e.g., Claude L\'evi-Strauss at 95 predicted as 45); (b)~apparent-age bias in celebrity datasets (CACD2000); (c)~mislabelling or incorrect detections during prediction (APPA-REAL, CACD2000); (d)~image quality and pose degradation (UTKFace, MORPH2).}
\label{fig:failures}
\end{figure*}

Fig.~\ref{fig:failures} shows the two worst predictions per public dataset (Intra setting), selected by absolute error. No National ID images are shown due to the privacy constraints described before.

%% ================================================================

\section{Lessons for National Identity Programmes}
\label{sec:use_cases}

In the Aadhaar enrollment pipeline, discrepancies between declared and estimated age arise from two sources: unintentional operator error (transposed digits, misread handwritten documents) and deliberate misrepresentation, with or without operator collusion. SwinAge addresses both by producing an independent estimate from the facial photograph alone. Logging the discrepancy for every  enrollment creates an audit trail; unusual patterns such as a specific operator or centre with a disproportionately high discrepancy rate can trigger targeted audits.

\textit{Threat model}: The estimate is independent of the documents but does not verify them. SwinAge catches data-entry error and misrepresentation where the applicant's appearance does not match the claim; it does not catch photo substitution, for which the compensating control is Aadhaar's existing biometric de-duplication. Robustness to appearance manipulation and degraded capture is not designed in; NIST reports strong image-quality dependence for this task~\cite{b33}.

\textit{Threshold-level performance}: With a whole-population MAE of  $\sim$2.9 years, performance differs across thresholds. The MAE of 0.75 for  ages 0--5 enables high-confidence detection of biometric bypass; the MAE of 2.07 for ages 15--20 gives sufficient precision at the adult-eligibility threshold; and the MAE of $\sim$3.76 for ages 55--65 is the largest of the three and drives the elevated $\geq$60 FRR in Table~\ref{tab:threshold}.

\textit{Operational volume}: At a $\pm$10-year discrepancy threshold, only 1.5\% of enrollments (4,126 of 283,515) are flagged for review, focusing operator attention on high-risk cases; $\pm$15 years reduces this to 0.2\% (629), while $\pm$5 years captures 19.5\% (55,214). The full pipeline runs at 187 images/s on one V100 GPU ($\sim$5.3~ms per image), bounded by I/O and CPU-side alignment rather than the model.

\subsection{Interpreting the Results}
\label{sec:lessons}

Three points in this work are generalizable to any program that weighs facial age estimation as an eligibility control.

% \textit{1) Report at the operating point, under a stated decision rule} (Sec.~\ref{sec:usecase_eval}). Aggregate MAE does not predict behaviour at an eligibility threshold: an MAE of 2.94 years coexists here with an FRR of 21.4\% at $\tau{=}60$ under an exact rule, and 11.0\% even with a band. State whether the margin was fixed on held-out data or, read off the test set; the latter is not comparable across systems and should be treated as an upper bound on operational performance.

\textit{1) Report at the operating point, under a stated decision rule} 
(Sec.~\ref{sec:usecase_eval}). Aggregate MAE does not predict behaviour at 
an eligibility threshold: an MAE of 2.94 years coexists here with an FRR of 
21.4\% at $\tau{=}60$ under an exact rule, and 11.0\% even with a band. 
State whether the margin was fixed on held-out data or read off the test set; 
the latter is not comparable across systems and should be treated as an upper 
bound on operational performance. When held-out calibration data are 
unavailable, $\delta{=}0.1\tau$ provides a model-agnostic fallback that 
aligns with per-threshold MAE for this model, though it does not guarantee 
a target FAR across all thresholds.

\textit{2) The label is the record under audit} (Sec.~\ref{sec:inhouse_data}). A model trained on declared dates of birth learns to agree with those records; reported accuracy is therefore agreement with the declared record, not with chronological age. Stratify evaluation by date-of-birth provenance, and where longitudinal enrollments exist, use self-consistency across a resident's own records as a free label-quality signal.

\textit{3) Sample for the subgroups you intend to audit} (Sec.~\ref{sec:age_gender}). Fairness should be evaluated at the decision level: a 0.3-year gender MAE gap does not itself characterize harm, whereas differences in false rejections or routing to manual review directly affect individuals. That audit is only possible if the test split supports it: ours had only 42 transgender images (26 subjects), too few to report on.

\section{Conclusion}
\label{sec:conclusion}

We presented SwinAge, a facial age estimation system designed for identity fraud prevention in the Aadhaar biometric enrollment pipeline. Using SwinFace with landmark-based warp affine alignment, we achieved an MAE of 2.94 on a 283K held-out test set from 1.75 million in-house images which is to our knowledge the first large-scale evaluation of age estimation on diverse demographics, predominantly of South Asian origin; with competitive performance on public benchmarks after finetuning. We tied age estimation to three concrete fraud use cases in the National ID system and proposed a deployment triage framework that flags cases where the predicted age crosses a threshold relative to the declared age, characterized its FAR/FRR under an exact rule, a fixed $\delta{=}0.1\tau$ band, and a calibrated band (FAR $\leq$1\% at all three thresholds; FRR 2.1\%, 0.8\%, 3.6\% under $0.1\tau$; FRR 3.0\%, 0.4\%, 11.0\% under the 
calibrated band).%characterized its FAR/FRR under an exact rule and a decision band (FAR 1.0\% at all three thresholds; FRR 3\%, 0.4\% and 11.0\%). 
The 60+ threshold remains the most challenging due to elevated MAE in this age range and the 75-year ceiling of the training corpus, and improving senior citizen age estimation is a focus of future work.

\paragraph*{Potential Societal Impacts} Facial age estimation at national scale raises privacy and fairness concerns. Demographic identifiers were removed (Sec.~\ref{sec:inhouse_data}), while facial images were accessed only for training and testing, under strict access control, for a limited duration, on-premises within the authority's data centre, and are never shared outside it. Processing is governed by \S23(2) and \S29(2) of the Aadhaar Act, 2016 and \S7(b) of the DPDP Act, 2023, with children's data (0--18) covered by guardian consent at enrollment per Regulation~6 of the Aadhaar (Enrolment and Update) Regulations, 2016; the work was approved and executed by the authority's technology centre. SwinAge is strictly decision-support: a flagged resident is told manual verification is required, may contest it before the quality check supervisor, and enrollment proceeds pending review, so no benefit is withheld on a flag alone. The discrepancy log is retained solely for audit, not as a standing risk score. The system should be audited regularly for demographic bias in decisions across age and skin tone.

Future work includes age progression as an auxiliary representation learning task to encode aging-aware features, leveraging a paired dataset of 2.17 million same-subject images at different ages, and label noise mitigation to improve estimation accuracy.

% --- COMMENTED FOR REVIEW FOR ANONYMITY ---- %
% --- Can be included in camera ready.
\paragraph*{Acknowledgements}
We acknowledge the open-source communities behind TensorFlow, PyTorch, FAISS and YOLOv8. We also thank Lokesh Kurre, Durgesh A. Sahane, Deepak K. Sial and Verish K. Meena for their valuable contributions.


\begin{thebibliography}{43}
\bibitem{b0} Unique Identification Authority of India (UIDAI), ``Aadhaar.'' [Online]. Available: \url{https://uidai.gov.in/}

\bibitem{b1} R. Rothe, R. Timofte, and L. Van Gool, ``DEX: Deep EXpectation of apparent age from a single image,'' in \textit{Proc. IEEE ICCV Workshops}, 2015.

\bibitem{b2} M. Kuprashevich and I. Tolstykh, ``MiVOLO: Multi-input Transformer for Age and Gender Estimation,'' in \textit{Proc. Int. Joint Conf. on Analysis of Images, Social Networks and Texts (AIST)}, 2023.

\bibitem{b3} Z. Liu, Y. Lin, Y. Cao, H. Hu, Y. Wei, Z. Zhang, S. Lin, and B. Guo, ``Swin Transformer: Hierarchical Vision Transformer using Shifted Windows,'' in \textit{Proc. IEEE/CVF ICCV}, 2021.

\bibitem{b4} L. Qin, M. Wang, C. Deng, K. Wang, X. Chen, J. Hu, and W. Deng, ``SwinFace: A Multi-task Transformer for Face Recognition, Expression Recognition, Age Estimation and Attribute Estimation,'' \textit{IEEE Transactions on Circuits and Systems for Video Technology}, vol. 34, no. 4, pp. 2223--2234, 2024.

\bibitem{b5} Y. Lin, J. Shen, Y. Wang, and M. Pantic, ``FP-Age: Leveraging Face Parsing Attention for Facial Age Estimation in the Wild,'' \textit{IEEE Transactions on Image Processing}, vol. 31, pp. 1779--1791, 2022.

\bibitem{b6} S. Moschoglou, A. Papaioannou, C. Sagonas, J. Deng, I. Kotsia, and S. Zafeiriou, ``AgeDB: The First Manually Collected, In-the-Wild Age Database,'' in \textit{Proc. IEEE CVPR Workshops (CVPRW)}, pp. 1997--2005, 2017.

\bibitem{b7} Z. Zhang, Y. Song, and H. Qi, ``Age Progression/Regression by Conditional Adversarial Autoencoder,'' in \textit{Proc. IEEE CVPR}, 2017.

\bibitem{b8} G. Panis, A. Lanitis, N. Tsapatsoulis, and T. F. Cootes, ``Overview of Research on Facial Ageing Using the FG-NET Ageing Database,'' \textit{IET Biometrics}, vol. 5, no. 2, pp. 37--46, 2016.

\bibitem{b9} E. Agustsson, R. Timofte, S. Escalera, X. Baro, I. Guyon, and R. Rothe, ``Apparent and real age estimation in still images with deep residual regressors on APPA-REAL database,'' in \textit{Proc. 12th IEEE Int. Conf. and Workshops on Automatic Face and Gesture Recognition (FG)}, pp. 87--94, 2017.

\bibitem{b10} B.-C. Chen, C.-S. Chen, and W. H. Hsu, ``Face Recognition and Retrieval Using Cross-Age Reference Coding With Cross-Age Celebrity Dataset,'' \textit{IEEE Transactions on Multimedia}, vol. 17, no. 6, pp. 804--815, 2015.

\bibitem{b11} K. Ricanek and T. Tesafaye, ``MORPH: A Longitudinal Image Database of Normal Adult Age-Progression,'' in \textit{Proc. IEEE Int. Conf. on Automatic Face and Gesture Recognition (FGR)}, pp. 341--345, 2006.

\bibitem{b12} G. Jocher, A. Chaurasia, and J. Qiu, ``Ultralytics YOLOv8,'' version 8.0.0, 2023. [Online]. Available: \url{https://github.com/ultralytics/ultralytics}

\bibitem{b13} S. Woo, J. Park, J.-Y. Lee, and I. S. Kweon, ``CBAM: Convolutional Block Attention Module,'' in \textit{Proc. European Conf. on Computer Vision (ECCV)}, pp. 3--19, 2018.

\bibitem{b14} J. Deng, J. Guo, N. Xue, and S. Zafeiriou, ``ArcFace: Additive Angular Margin Loss for Deep Face Recognition,'' in \textit{Proc. IEEE/CVF CVPR}, 2019.

\bibitem{b15} R. Ranjan, S. Zhou, and R. Chellappa, ``Unconstrained Age Estimation with Deep Convolutional Neural Networks,'' in \textit{Proc. IEEE Int. Conf. on Computer Vision Workshops (ICCVW)}, pp. 1--9, 2015.

\bibitem{b16} I. Loshchilov and F. Hutter, ``Decoupled Weight Decay Regularization,'' in \textit{Proc. Int. Conf. on Learning Representations (ICLR)}, 2019.

\bibitem{b17} Parliament of India, ``The Aadhaar (Targeted Delivery of Financial and Other Subsidies, Benefits and Services) Act,'' 2016. [Online]. Available: \url{https://uidai.gov.in/legal}

\bibitem{b18} Parliament of India, ``The Digital Personal Data Protection Act,'' 2023. [Online]. Available: \url{https://dpdpa.com/DPDPA_2023_official.pdf}

\bibitem{b19} P. Grd, E. Bar\v{c}i\'{c}, I. Tomi\v{c}i\'{c}, and B. Okre\v{s}a \DJ{}uri\'{c}, ``Analysing the Impact of Gender Classification on Age Estimation,'' in \textit{Proc. 12th European Conf. on Software Engineering and Information Technology (ECSEIT)}, pp. 53--59, 2023.

\bibitem{b20} J. C. S. Jacques Junior, C. Ozcinar, M. Marjanovic, X. Bar{\'o}, G. Anbarjafari, and S. Escalera, ``On the effect of age perception biases for real age regression,'' in \textit{Proc. 14th IEEE Int. Conf. on Automatic Face and Gesture Recognition (FG)}, pp. 1--7, 2019.

\bibitem{b22} J. Paplham and V. Franc, ``A Call to Reflect on Evaluation Practices for Age Estimation: Comparative Analysis of the State-of-the-Art and a Unified Benchmark,'' in \textit{Proc. IEEE/CVF CVPR}, pp. 9858--9867, 2024.

\bibitem{b23} C. Yen, J. Ding, and K. Hu, ``AGMixer: Age Estimation Using Gender Feature and Improved Ordinal Loss,'' in \textit{Proc. IEEE Int. Symp. on Circuits and Systems (ISCAS)}, pp. 1--5, 2025.

\bibitem{b24} B. Wu, Z. Ai, J. Jiang, C. Zhu, and S. Xu, ``Stage-wise Adaptive Label Distribution for Facial Age Estimation,'' \textit{arXiv preprint arXiv:2509.00450}, 2025.

\bibitem{b25} Comptroller and Auditor General of India, ``Performance Audit of National Social Assistance Programme in Ministry of Rural Development,'' Report No. 10 of 2023, New Delhi, India, 2023.

\bibitem{b26} World Bank, ``Thailand: Social Protection and Labor Assessment,'' World Bank Report, 2012.

\bibitem{b27} NVIDIA, ``Nemotron-Nano-12B-v2-VL,'' Hugging Face model, 2025. [Online]. Available: \url{https://huggingface.co/nvidia/nemotron-nano-12b-v2-vl}

\bibitem{b28} Qwen Team, ``Qwen3-VL-235B-A22B-Instruct,'' Hugging Face model, 2025. [Online]. Available: \url{https://huggingface.co/Qwen/Qwen3-VL-235B-A22B-Instruct-FP8}

\bibitem{b29} Google, ``Gemma-4-31B-IT,'' Hugging Face model, 2025. [Online]. Available: \url{https://huggingface.co/google/gemma-4-31B-it}

\bibitem{b30} Unique Identification Authority of India, ``UIDAI Annual Report 2024--25,'' Ministry of Electronics and Information Technology, Government of India, New Delhi, 2025. [Online]. Available: \url{https://uidai.gov.in/en/annual-reports}

\bibitem{b31} M. Kuprashevich, G. Alekseenko, and I. Tolstykh, ``Beyond Specialization: Assessing the Capabilities of MLLMs in Age and Gender Estimation,'' arXiv preprint arXiv:2403.02302, 2024.

\bibitem{b32} Z. Niu, M. Zhou, L. Wang, X. Gao, and G. Hua, ``Ordinal Regression with Multiple Output CNN for Age Estimation,'' in \textit{Proc. IEEE Conf. on Computer Vision and Pattern Recognition (CVPR)}, pp. 4920--4928, 2016.

\bibitem{b33} K. Hanaoka, M. Ngan, J. Yang, G. W. Quinn, A. Hom, and P. Grother, ``Face Analysis Technology Evaluation: Age Estimation and Verification,'' NIST Internal Report NIST IR 8525, National Institute of Standards and Technology, Gaithersburg, MD, 2024. doi: 10.6028/NIST.IR.8525

\bibitem{b34} Y. Lin, J. Shen, Y. Wang, and M. Pantic, ``RoI Tanh-polar transformer network for face parsing in the wild,'' \textit{Image and Vision Computing}, vol. 112, 104190, 2021.

\bibitem{b35} J. T. Barron, ``A General and Adaptive Robust Loss Function,'' in \textit{Proc. IEEE/CVF CVPR}, pp. 4331--4339, 2019.

\bibitem{b36} X. Geng, C. Yin, and Z.-H. Zhou, ``Facial Age Estimation by Learning from Label Distributions,'' \textit{IEEE Transactions on Pattern Analysis and Machine Intelligence}, vol. 35, no. 10, pp. 2401--2412, 2013.

\bibitem{b37} B.-B. Gao, C. Xing, C.-W. Xie, J. Wu, and X. Geng, ``Deep Label Distribution Learning with Label Ambiguity,'' \textit{IEEE Transactions on Image Processing}, vol. 26, no. 6, pp. 2825--2838, 2017.

\bibitem{b38} W. Cao, V. Mirjalili, and S. Raschka, ``Rank consistent ordinal regression for neural networks with application to age estimation,'' \textit{Pattern Recognition Letters}, vol. 140, pp. 325--331, 2020.

\bibitem{b39} H. Pan, H. Han, S. Shan, and X. Chen, ``Mean-Variance Loss for Deep Age Estimation from a Face,'' in \textit{Proc. IEEE/CVF CVPR}, pp. 5285--5294, 2018.

\bibitem{b40} A. Puc, V. \v{S}truc, and K. Grm, ``Analysis of Race and Gender Bias in Deep Age Estimation Models,'' in \textit{Proc. 28th European Signal Processing Conf. (EUSIPCO)}, pp. 830--834, 2020.

\bibitem{b41} Unique Identification Authority of India, ``Aadhaar (Enrolment and Update) Regulations, 2016,'' as amended. [Online]. Available: \url{https://uidai.gov.in/en/about-uidai/legal-framework/regulations}

\bibitem{b42} Ministry of Electronics and Information Technology, Government of India, ``Aadhaar authentication surges past 2,707 Cr in 2024--25; Aadhaar e-KYC transactions close to 45 Cr in March,'' Press Information Bureau, Press Release ID 2124910, 28 April 2025. [Online]. Available: \url{https://www.pib.gov.in/PressReleasePage.aspx?PRID=2124910}

\bibitem{b43} ISO/IEC, ``Information security, cybersecurity and privacy protection --- Age assurance systems --- Part 1: Framework,'' ISO/IEC 27566-1:2025, International Organization for Standardization, Geneva, 2025.

\bibitem{b44} E. D. Cubuk, B. Zoph, J. Shlens, and Q. V. Le, ``RandAugment: Practical Automated Data Augmentation with a Reduced Search Space,'' in \textit{Proc. Advances in Neural Information Processing Systems (NeurIPS)}, 2020.

\bibitem{b45} Z. Zhong, L. Zheng, G. Kang, S. Li, and Y. Yang, ``Random Erasing Data Augmentation,'' in \textit{Proc. AAAI Conference on Artificial Intelligence}, vol. 34, no. 07, pp. 13001--13008, 2020.

\end{thebibliography}
\end{document}